\documentclass[sigconf]{acmart}
\AtBeginDocument{%
  }

\setcopyright{acmlicensed}
\copyrightyear{2018}
\acmYear{2018}
\acmDOI{XXXXXXX.XXXXXXX}
\acmConference[Conference acronym 'XX]{}{June 03--05,
  2018}{Woodstock, NY}
\acmISBN{978-1-4503-XXXX-X/2018/06}

\usepackage{multirow}
\usepackage{graphicx}
\usepackage{subcaption}
\copyrightyear{2025}
\acmYear{2025}
\setcopyright{acmlicensed}\acmConference[MM '25]{Proceedings of the 33rd ACM International Conference on Multimedia}{October 27--31, 2025}{Dublin, Ireland}
\acmBooktitle{Proceedings of the 33rd ACM International Conference on Multimedia (MM '25), October 27--31, 2025, Dublin, Ireland}
\acmDOI{10.1145/3746027.3754587}
\acmISBN{979-8-4007-2035-2/2025/10}
\begin{document}

\title{FedDEAP: Adaptive Dual-Prompt Tuning for Multi-Domain Federated Learning}

\author{Yubin Zheng}
\affiliation{%
  \institution{Shanghai Jiao Tong University}
  \city{Shanghai}
  \country{China}
}
\email{zybhk21@sjtu.edu.cn}

\author{Pak-Hei Yeung}
\affiliation{%
  \institution{Nanyang Technological University}
  \country{Singapore}}
\email{pakhei.yeung@ntu.edu.sg}

\author{Jing Xia}
\affiliation{%
  \institution{Nanyang Technological University}
  \country{Singapore}
}
\email{xiajing0904@gmail.com}

\author{Tianjie Ju}
\affiliation{%
  \institution{Shanghai Jiao Tong University}
  \city{Shanghai}
  \country{China}
}
\email{jometeorie@sjtu.edu.cn}

\author{Peng Tang}
\affiliation{%
  \institution{Shanghai Jiao Tong University}
  \city{Shanghai}
  \country{China}
}
\email{tangpeng@sjtu.edu.cn}

\author{Weidong Qiu}
\authornote{Corresponding author}
\affiliation{%
  \institution{Shanghai Jiao Tong University}
  \city{Shanghai}
  \country{China}
}
\email{qiuwd@sjtu.edu.cn}

\author{Jagath C. Rajapakse}
\affiliation{%
  \institution{Nanyang Technological University}
  \country{Singapore}
}
\email{asjagath@ntu.edu.sg}

\renewcommand{\shortauthors}{Yubin Zheng et al.}

\begin{abstract}
 Federated learning (FL) enables multiple clients to collaboratively train machine learning models without exposing local data, balancing performance and privacy. However, domain shift and label heterogeneity across clients often hinder the generalization of the aggregated global model. Recently, large-scale vision-language models like CLIP have shown strong zero-shot classification capabilities, raising the question of how to effectively fine-tune CLIP across domains in a federated setting. In this work, we propose an adaptive federated prompt tuning framework, FedDEAP, to enhance CLIP’s generalization in multi-domain scenarios. Our method includes the following three key components: (1) To mitigate the loss of domain-specific information caused by label-supervised tuning, we disentangle semantic and domain-specific features in images by using semantic and domain transformation networks with unbiased mappings; (2) To preserve domain-specific knowledge during global prompt aggregation, we introduce a dual-prompt design with a global semantic prompt and a local domain prompt to balance shared and personalized information; (3) To maximize the inclusion of semantic and domain information from images in the generated text features, we align textual and visual representations under the two learned transformations to preserve semantic and domain consistency. Theoretical analysis and extensive experiments on four datasets demonstrate the effectiveness of our method in enhancing the generalization of CLIP for federated image recognition across multiple domains.
\end{abstract}





\maketitle

\section{Introduction}
The rapid advancement of deep learning has been largely driven by large-scale models trained on massive datasets. However, due to the increasing concerns over data privacy, aggregating data from various sources to construct a centralized dataset has become impractical. Federated learning (FL) \cite{mcmahan2017communication} is a distributed machine learning paradigm that enables multiple clients to collaboratively train a model without sharing their local data. Each client trains a model using its private dataset and periodically transmits the locally updated model to a central server. The server aggregates these local models and redistributes the refined global model to the clients, facilitating an iterative optimization process. The decentralized FL framework effectively balances model performance and privacy preservation. 

FL has been widely applied in real-world scenarios such as smart healthcare \cite{gong2021ensemble, li2020multi, zheng2024federated, wu2021federated}, autonomous driving \cite{su2023cross, li2021privacy}, and the financial sector \cite{liu2021fate}. However, one of the critical challenges in FL is data heterogeneity, which arises when client data originates from different domains and exhibits label distribution discrepancies \cite{li2020fedprox, yao2022enhancing, yang2021covid, liu2021feddg}. As a result, the gradient optimization directions of locally trained models differ, and aggregating these divergent local updates at the central server can hinder global model convergence and degrade its performance. Therefore, enhancing the generalization ability of the global model across heterogeneous domains remains a key research challenge in FL.

Vision-language foundation models, such as CLIP \cite{radford2021clip}, are pre-trained on large-scale image-text pairs through contrastive learning that aligns visual and textual feature spaces \cite{li2022blip, wang2022medclip}. These models have shown remarkable zero-shot classification performance in various downstream tasks due to their strong representation learning capabilities. Given the public availability of CLIP and its impressive few-shot learning abilities, CLIP becomes a strong candidate in FL applications. This motivates us to explore how clients can collaboratively perform efficient federated prompt tuning \cite{lester2021prompt} on CLIP, leveraging its powerful representation capacity to jointly learn image classification tasks across multiple domains.

However, heterogeneous feature and label distributions across domains pose significant challenges for federated prompt tuning, as directly aggregating prompt parameters from different clients can compromise the generalization performance of the global model due to the domain shift and label heterogeneity. This raises a key question: "\textbf{How can we optimize the prompts to ensure that the generated textual features can contain the most similar semantic and domain information with image features?}"

To address this issue, we introduce \textbf{FedDEAP}, a \textit{Federated framework for Dual-prompt and ETF-alignment Adaptive Prompt tuning} tailored to the CLIP model. Our approach aims to balance global knowledge sharing with the preservation of local domain-specific features. To achieve this, we propose using two joint prompts including a global semantic prompt shared across different domains and a personalized domain-specific prompt trained locally. The global semantic prompt facilitates capturing global semantic features across all domains, while the domain-specific prompt ensures the model retains essential local domain information. Additionally, we introduce unbiased transformation networks constrained by Equiangular Tight Framework (ETF) \cite{papyan2020etf} to decouple the semantic and domain spaces within images and align the learned global and local prompts with image features in semantic and domain spaces, respectively. The global semantic alignment via the unbiased semantic transformation network constrains the semantic bias of the global prompts during local training, thereby mitigating performance degradation caused by label heterogeneity across clients. Meanwhile, the domain alignment through the unbiased domain transformation network enhances the local domain prompts to capture more discriminative domain-specific features, thus improving cross-domain generalization. This strategic separation of prompts mitigates the loss of domain-specific knowledge during federated training, ultimately leading to improved generalization and robust performance across multiple domains. 

Compared to existing baselines, our proposed FedDEAP achieves state-of-the-art classification performance on three natural image datasets and one medical image dataset under different heterogeneous settings. Additionally, our method achieves faster inference, offering both high efficiency and superior performance. Our main contributions are summarized as follows:

\begin{itemize}
    \item A dual-prompt strategy is proposed, consisting of a global semantic prompt and a local domain prompt, which are refined by unbiased transformation networks to align with image features in both semantic and domain spaces.
    \item Extensive experiments indicate that our approach achieves the state-of-the-art performance across three natural image datasets and one medical image dataset compared to strong baselines.
    \item Our theoretical analysis and detailed ablation studies further confirm that the proposed dual-prompt and alignment strategies effectively preserve semantic and domain information of images in the learned prompts.
\end{itemize}

\begin{figure*}[h]
  \centering
  \includegraphics[width=\linewidth]{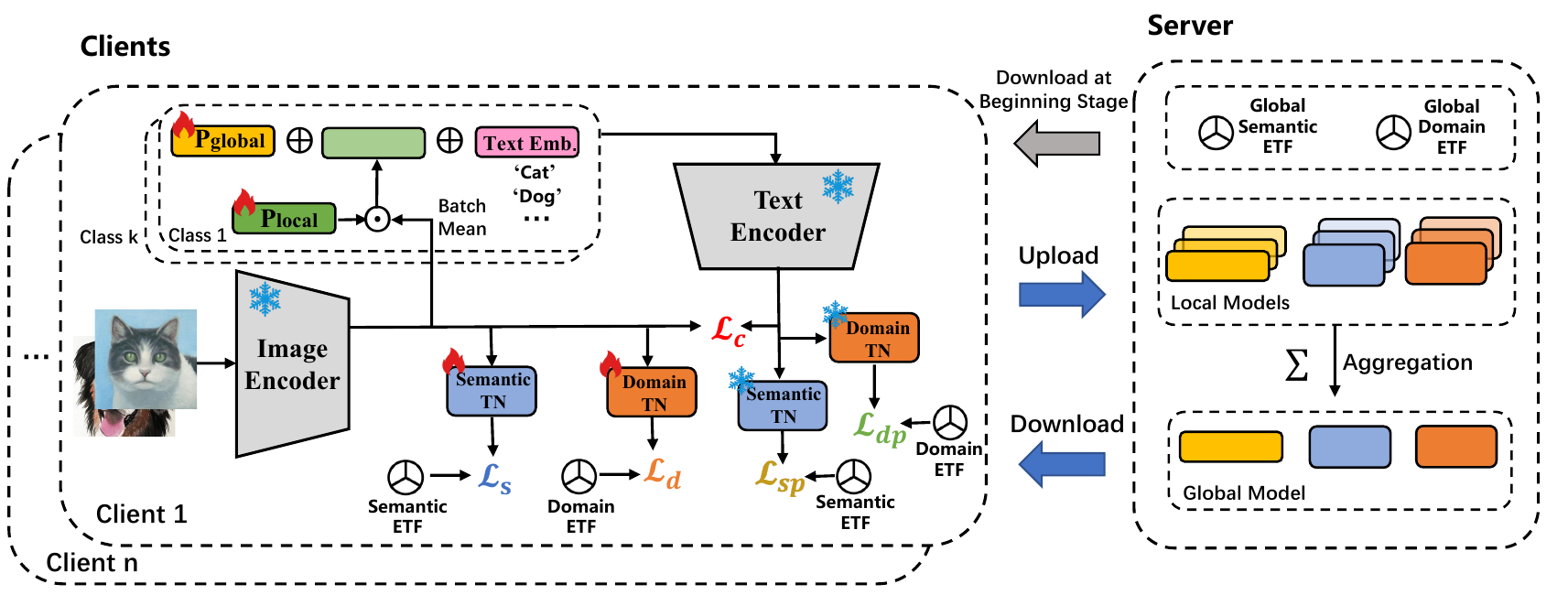}
  \caption{The framework of the proposed FedDEAP method. Each client decouples semantic and domain features from images using transformation networks (TNs) guided by global semantic and domain ETF, and aligns the prompt with the image in both feature spaces. The global prompt and TNs are aggregated on the server at the end of each round.}
\end{figure*}

\section{Related Work}
\subsection{Federated Learning}
Google first proposed the concept of utilizing user devices for distributed model training and introduced the FedAvg algorithm \cite{mcmahan2017communication}, which has become a foundational approach in FL. However, FL faces challenges due to data heterogeneity across clients, including differences in feature and label distributions, which slow down convergence and degrade generalization. To address this, numerous studies have proposed improved FL algorithms \cite{li2021moon, karimireddy2020scaffold, zhu2021fedgen, wang2020fednova}. For example, FedProx \cite{li2020fedprox} enhances FedAvg by introducing a regularization term to mitigate divergence between local and global models, thus improving stability and convergence. 

Personalized federated learning \cite{fallah2020mocha, t2020envelopes, jeong2018FAG} has recently gained attention. It aims to balance cross-client knowledge sharing with local model optimization to better adapt to heterogeneous data and models. Some typical approaches include knowledge distillation \cite{li2019fedmd, wu2022NC, gong2022fedkd}, which transfers global knowledge to enhance local model performance, and parameter decoupling \cite{liang2020LG-Fedavg,zhu2021fedgen, tan2022fedpcl} that separates shared and personalized components to enhance local adaptability.

With the rise of large-scale pre-trained models achieving remarkable performance across various tasks, many studies have explored integrating such models into FL \cite{yang2024stabledf, saha2024fedpia, zhang2024upload, tan2022fedpcl}. For instance, FedDEO \cite{yang2024feddeo} trains local Stable Diffusion description vectors related to each client’s data and generates synthetic data on the server to improve global model training. These approaches demonstrate the potential of leveraging foundation models to advance federated learning.

\subsection{CLIP and Prompt Tuning}
CLIP \cite{radford2021clip} is a vision-language model that leverages contrastive learning to align visual and textual feature spaces, pre-trained on large-scale image-text pairs. It demonstrates strong zero-shot classification capabilities across various computer vision tasks. By leveraging CLIP, traditional image classification can be reformulated as a matching problem between image features and text features of different categories.

Although CLIP already exhibits powerful image-text alignment, its performance can be further enhanced through fine-tuning on downstream datasets. Prompt tuning \cite{lester2021prompt, li2021prefix} is an efficient parameter-efficient fine-tuning (PEFT) method \cite{houlsby2019adapter, hu2022lora} for large pre-trained models, which adapts models to specific tasks by optimizing learnable text tokens instead of updating the entire model. This approach has been first successfully applied to CLIP, with CoOp \cite{zhou2022coop} introducing prompt tuning for CLIP by appending trainable prompts to category text descriptions. CoCoOp \cite{zhou2022cocoop} extends this approach by employing a meta-network to map image features into a meta-token and integrating this token into the original trainable prompts to improve CLIP’s generalization to unseen categories.

\subsection{Federated Fine-tuning for CLIP}
Research on federated fine-tuning of CLIP remains relatively limited. Due to communication efficiency considerations in FL, existing works adopt PEFT methods to collaboratively fine-tune CLIP. For example, PromptFL \cite{guo2023promptfl} builds upon CoOp by aggregating local prompts to achieve joint fine-tuning of CLIP across clients. FedCLIP \cite{lu2023fedclip} introduces an adapter after the image encoder to fine-tune CLIP in multi-domain FL scenarios. FACMIC \cite{wu2024facmic} improves upon FedCLIP by incorporating an inter-domain regularization term to better handle varying data distributions across clients. More recently, FedAPT\cite{su2024fedapt} has further enhanced federated prompt tuning for CLIP by incorporating client-assigned keys into the meta-prompt, enabling the global model to dynamically generate specific prompts of each client. However, some of these methods overlook the data heterogeneity problem and possess large inference overhead. In this work, we focus on textual prompt tuning methods for CLIP. Our proposed FedDEAP is capable of learning domain-adaptive prompts for each client, effectively addressing both domain shift and label heterogeneity, while maintaining fast inference efficiency.

\subsection{Methods}
The overall framework of the proposed FedDEAP is illustrated in Figure 1. We first define the problem of Federated Prompt Tuning for CLIP in multi-domain scenarios in Section 3.1. In Section 3.2, we describe how our framework utilizes the Equiangular Tight Framework (ETF) to derive unbiased semantic and domain representations of images. Section 3.3 presents the training of global semantic prompts and local domain prompts with the assistance of semantic and domain transformation networks. Finally in Section 3.4, we theoretically show that our framework improves the mutual information lower bound between textual and visual features at both semantic and domain levels.

\subsection{Problem Definition}
Consider a federated learning scenario where multiple clients, such as medical institutions, exhibit domain shift in data. We assume that all clients have access to a pre-trained CLIP. Instead of fine-tuning the entire model, clients employ federated prompt tuning while keeping the pre-trained CLIP model frozen.

At the client level, the prompt tuning procedure follows the standard CLIP-based prompt tuning method. It adapts the CLIP model to downstream tasks by constructing and optimizing learnable prompt templates. Let's consider a client with a local dataset $D=(x,y)$, where $x$ represents the image data and $y$ corresponds to its class label. For each class, the client constructs a learnable prompt $\text{p}=[u_1, u_2,...,u_l]$, consisting of $l$ learnable vectors, each with the size of a token embedding. The prompt is then concatenated with the embedding of a textual prompt, such as "a photo of a dog", to construct the input $(\text{p};\text{t})$ for the CLIP text encoder $\mathcal{T}$. Let $\mathcal{I}$ be the CLIP image encoder, the prompt tuning goal of the local CLIP model is to maximize the probability:
\begin{equation}
    p(y|x)=\frac{\exp(\cos(\mathcal{T}(\text{p}^y;\text{t}^y), \mathcal{I}(x))/\tau)}{\sum_{k=1}^{K}\exp(\cos(\mathcal{T}(\text{p}^k;\text{t}^k), \mathcal{I}(x))/\tau)}
\end{equation}
where $K$ represents the total number of classes, $\tau$ denotes the temperature parameter.

In a federated learning setting, a fundamental approach to prompt tuning CLIP is to conduct local prompt aggregation through the FedAvg method. Each client uploads its locally trained prompts to a central server. The server aggregates the local prompts using the FedAvg method and subsequently distributes the updated global prompt to all clients, enabling iterative optimization. Consequently, the objective of the federated prompt tuning on CLIP is to minimize the following loss:
\begin{equation}
    \mathcal{L}=\frac{1}{N}\sum_{n=1}^N\mathbb{E}_{(x,y)\in D_n}\ell(\text{p}_g,x,y)
\end{equation}
where $\ell$ represents the classification loss on local data using the global prompt $\text{p}_g$, $N$ is the total number of the clients, and $D_n$ denotes the local data of client $n$.

\subsection{Semantic and Domain Transformation Networks Training with ETF Classifier}
In Federated Prompt Tuning for the CLIP model, the fine-tuning loss based on contrastive loss with class labels can cause prompts to prioritize semantic features over domain characteristics. Moreover, the non-IID class distribution across clients can degrade semantic recognition performance due to the aggregation of discrepant local updates in federated learning. To address these challenges, we propose Semantic and Domain Transformation Networks, which apply non-linear transformations to CLIP image representations, enabling the learning of global semantic features while preserving local domain characteristics.

To learn an unbiased transformation of semantic and domain features, we draw inspiration from FedETF \cite{li2023fedetf} and employ an Equiangular Tight Frame (ETF) classifier to constrain their representations. The ETF classifier consists of class prototypes arranged in an equiangular tight frame, where each pair of prototypes exhibits the same pairwise cosine similarity. 

Taking the semantic ETF as an example, we formally define a set of semantic ETF prototype vectors $V_s=\{v_s^1,v_s^2,...,v_s^K\}$, where $V_s=\sqrt{\frac{K}{K-1}}U(I_K-\frac{1}{K}1_K1_K^{\top}) \in \mathbb{R}^{M\times K}$. Here, $U$ is a dimensional transformation matrix that satisfies $U \in \mathbb{R}^{M\times K}$ and $U^TU=I_K$. For any $v_s^k$ in the ETF prototypes, it holds that $||v_s^k||^2=1$. Moreover, for any $k_1,k_2 \in [K]$ with $k_1 \neq k_2$, the cosine similarity satisfies $\cos(v_s^{k_1},v_s^{k_2})=-\frac{1}{K-1}$. This property of the ETF classifier maximizes inter-class separability while ensuring intra-class compactness, thereby guaranteeing the unbiasedness of the feature representations learned by the semantic and domain transformation networks across different clients.

The server first initializes the semantic ETF $V_s = \{v_s^1, \dots, v_s^K\} \in \mathbb{R}^{M \times K}$ and the domain ETF $V_d = \{v_d^1, \dots, v_d^N\} \in \mathbb{R}^{M \times N}$, where $ K $ denotes the number of semantic classes and $N$ represents the number of clients (domains).  
These initialized prototypes are then distributed to all participating clients.

For the $n$-th client, the local image data $(x, y)$ is utilized to train the semantic transformation network $\Phi_s^n$ and the domain transformation network $\Phi_d^n$. The training objective of $\Phi_s^n$ is to minimize the angular discrepancy between the transformed representation $\Phi_s^n(\mathcal{I}(x))$ and the corresponding prototype vector $v_s^y$ in the semantic ETF $V_s$. Thus, the optimization objective for the semantic transformation network is defined as:
\begin{equation}
    \mathcal{L}_s = \mathbb{E}_{(x,y) \sim D_n} 
\left[ -\log \frac{\exp({\cos({\Phi_s^n(\mathcal{I}(x))},v_s^y) / \tau})}
{\sum_{k=1}^{K} \exp({\cos(\Phi_s^n(\mathcal{I}(x))}, v_s^k) / \tau)} \right]
\end{equation}

Similarly, the training objective for the domain transformation network is to minimize the angular discrepancy between the transformed representation $\Phi_d^n(\mathcal{I}(x))$ and the corresponding prototype vector $ v_d^n $ in the domain ETF $ V_d $, which is formally expressed as:
\begin{equation}
    \mathcal{L}_d = \mathbb{E}_{(x,y) \sim D_n} 
\left[ -\log \frac{\exp({\cos{(\Phi_d^n(\mathcal{I}(x))}, v_d^n) / \tau})}
{\sum_{i=1}^{N} \exp({\cos(\Phi_d^n(\mathcal{I}(x))}, v_d^i) / \tau)} \right]
\end{equation}

During training, the CLIP encoder and ETF parameters remain frozen, and only the parameters of the two transformation networks are updated. Through training, the global semantic transformation network and the domain transformation network can effectively decouple semantic and domain-specific feature representations from the pre-trained CLIP embeddings across different clients. 

\subsection{Global Shared Semantic and Local Personalized Domain Prompts Training}
In the multi-domain federated learning scenario, fine-tuned prompts across different domains often suffer from degraded generalization ability when aggregated globally. This issue arises due to the heterogeneous nature of data across domains. To address this challenge, we introduce a global shared semantic prompt $\text{p}_s\in \mathbb{R}^{K\times L \times D}$ and a local personalized domain-specific prompt $\text{p}_d\in \mathbb{R}^{K\times L \times D}$, where $L$ denotes the number of prompt vectors and $D$ represents the dimension of the vector. The global prompt is designed to capture shared semantic representations across domains, while the local prompt is tailored to domain-specific features.

Local images within each client (domain) share a consistent style, even if they belong to different categories. To enhance the local prompt’s ability to encode domain-specific characteristics, we integrate image information into the local prompt by performing the Hadamard product between the local prompt and the mean feature embeddings extracted from a batch of images via the CLIP image encoder. Subsequently, for each client $n$, a locally updated copy of the global semantic prompt $\text{p}_s^n$ is concatenated with the personalized domain prompt $\text{p}_d^n$ and the text embeddings $E_\text{text}$ to form the input to the CLIP text encoder:
\begin{equation}
    \text{p}_d^n=\text{Batch\_Mean}(\mathcal{I}(x_\text{batch}))\odot\text{p}_d^n
\end{equation}
\begin{equation}
    E = \text{p}_s^n \oplus \text{p}_d^n \oplus E_\text{text}
\end{equation}
where $\odot$ denotes the element-wise product, $\oplus$ denotes the concatenated operation, and $E_\text{text}$ represents the embedding of the text.

During the training process of the global semantic prompt $\text{p}_s^n$ in each client $n$, the primary optimization objective is to minimize the contrastive loss $L_c$ between the text and image features:
\begin{equation}
    \mathcal{L}_c = \mathbb{E}_{(x,y) \sim D_n} \left[ -\log \frac{\exp(\cos(\mathcal{T}(E_y),\mathcal{I}(x))/\tau)}
    {\sum_{k=1}^{K} \exp(\cos(\mathcal{T}(E_k),\mathcal{I}(x))/\tau)} \right]
\end{equation}
Building on this, to ensure that the same class across different clients (domains) learns similar semantic features, the generated text features are projected through the semantic transformation network $\Phi_s^n$. The transformed features are then aligned with the class prototypes in the global semantic ETF $V_s$. Formally, this semantic similarity loss is defined as:
\begin{equation}
    \mathcal{L}_{sp} = \mathbb{E}_{y \sim K} \left[ -\log \frac{\exp(\cos(\Phi_s^n(\mathcal{T}(E_y)), v_s^y)/\tau)}
    {\sum_{k=1}^{K} \exp(\cos(\Phi_s^n(\mathcal{T}(E_y)), v_s^k)/\tau)} \right]
\end{equation}
By integrating both loss functions, the training objective of the global semantic prompt $\text{p}_s^n$ is formulated as:
\begin{equation}
    \mathcal{L}_{pg}=\mathcal{L}_{c}+\lambda\mathcal{L}_{sp}
\end{equation}
We employ a local personalized prompt $\text{p}_d^n$ for each client $n$ to capture the unique domain-specific features of local data. To prevent the personalized prompt from degrading the fine-tuning performance of the local CLIP model, we optimize it using the same contrastive loss $\mathcal{L}_c$ in the fine-tuning process. Meanwhile, to effectively learn domain-specific representations, the text feature vectors are projected through the domain transformation network $\Phi_d^n$. The transformed features are then aligned with the class prototypes of the corresponding domain in the domain-specific ETF $V_d$. Formally, this domain similarity loss is defined as:
\begin{equation}
    \mathcal{L}_{dp} = \mathbb{E}_{y \sim K} \left[ -\log \frac{\exp(\cos(\Phi_d^n(\mathcal{T}(E_y)), v_d^n)/\tau)}
    {\sum_{i=1}^{N} \exp(\cos(\Phi_d^n(\mathcal{T}(E_y)), v_d^i)/\tau)} \right]
\end{equation}
Consequently, the training goal of the local domain prompt $\text{p}_d^n$ is:
\begin{equation}
    \mathcal{L}_{pl}=\mathcal{L}_{c}+\eta\mathcal{L}_{dp}
\end{equation}

During the training of semantic and domain prompts, the parameters of two transformation networks remain frozen. The global semantic prompt $\text{p}_s^n$ of each client $n$ is aggregated to learn shared semantics across clients, while the local personalized prompt $\text{p}_d^n$ is trained locally without aggregation, focusing on domain-specific features. The semantic and domain transformation networks, $\Phi_s^n$ and $\Phi_d^n$, are also aggregated in the server each round:
\begin{equation}
    p_s^g = \frac{1}{N}\sum_{n=1}^N p_s^n, \quad \Phi_s^g = \frac{1}{N}\sum_{n=1}^N \Phi_s^n, \quad \Phi_d^g = \frac{1}{N}\sum_{n=1}^N \Phi_d^n
\end{equation}
By jointly optimizing the global semantic prompt and the local domain prompt, the text features incorporate both semantic cues and domain-specific adaptations. This design can enhance the model’s ability to generalize across diverse domains while maintaining strong performance in global semantic classification.

\subsection{Mutual Information Preservation Analysis}
In this section, we provide a theoretical analysis to demonstrate that the mutual information in domain space $I(r_t, r_i|k, d)$ and in semantic space $I(r_t, r_i|k, s)$ between the text feature representation $r_t$, obtained from the text encoder, and the image feature representation $r_i$, obtained from the image encoder, has a significant lower bound within the same class $k$. This implies that the shared semantic or domain-specific information of the two representations reaches a sufficiently high level.

Let $r_t, r_i \in \mathbb{R}^d$ be the feature representations belonging to the same class $k$. Taking the semantic transformation mapping as an example:
\begin{equation}
    I(r_t;r_i|k, s) \approx I(\Phi_s(r_t);\Phi_s(r_i)|k) + \text{const}
\end{equation}
Due to the equiangular tight frame (ETF) property, in class $k$:
\begin{equation}
    P(||\Phi_s(r_t)-\Phi_s(r_i)||\leq \delta | k) \geq \gamma
\end{equation}
where $\delta$ is a small value. $\gamma$ exhibits a monotonic increase as 
$\delta$ increases. Specifically, when $\delta = \sqrt{2 - \sqrt{\frac{2K - 4}{K - 1}}}$, we have $\gamma \to 1$ (see Appendix A for details), where $K$ represents the number of classes in the ETF.
The mutual information $I(\Phi_s(r_t);\Phi_s(r_i)|k)$ can also be written as:
\begin{equation}
    I(\Phi_s(r_t);\Phi_s(r_i)|k) = H(\Phi_s(r_t)|k) - H(\Phi_s(r_t)|\Phi_s(r_i), k)
\end{equation}
When $\Phi_s(r_t)$ and $\Phi_s(r_i)$ are very close, i.e., $||\Phi_s(r_t) - \Phi_s(r_i)|| \leq \delta$, the shared information between them is significantly large, and the upper bound of the conditional entropy is:
\begin{equation}
    H(\Phi_s(r_t)|\Phi_s(r_i), k) \leq \log V \approx \alpha d\log \delta + \text{const}
\end{equation}
where $V$ is the volume of a sphere with radius $\delta$ in a $d$-dimensional space and $\alpha$ is a very small factor. Since $\alpha d \propto K - 1$ due to the effective feature space under the ETF constraint and $H(\Phi_s(r_t) | k) > B$ (see Appendix B for details):
\begin{align}
    B = \log \left[ e + (K - 1) e^{-1/(K - 1)} \right]
- \frac{e - e^{-1/(K - 1)}}{e + (K - 1) e^{-1/(K - 1)}}
\end{align}
Then we obtain:
\begin{align}
    I(r_t; r_i \mid k, s) &\approx I(\Phi_s(r_t); \Phi_s(r_i) \mid k) + \text{const} \notag \\
                        &\approx H(\Phi_s(r_t) \mid k) - H(\Phi_s(r_t) \mid \Phi_s(r_i), k) + \text{const} \notag \\
                        & > B - \alpha d \log \delta + \text{const} \notag \\
                        & > B - \frac{K-1}{2}\log{(2-\sqrt{\frac{2K-4}{K-1}})} + \text{const}
\end{align}

The theoretical result supports that ETF-constrained transformations enable prompt-tuned features to retain high mutual information with image embeddings in both semantic and domain spaces, thus enhancing generalization in multi-domain image recognition.

\begin{table*}[h]
    \centering
    \caption{Comparison of classification accuracy between the FedDEAP and baselines on PACS, DomainNet, and Office datasets, evaluated across individual domains and their averages.}
    \renewcommand{\arraystretch}{1.05}
    \resizebox{\textwidth}{!}{
    \begin{tabular}{l|ccccc||ccccc||ccccc}
        \toprule
        \multirow{2}{*}{Method} & \multicolumn{5}{c||}{PACS} & \multicolumn{5}{c||}{DomainNet} & \multicolumn{5}{c}{Office}\\
        \cline{2-16}
         & a & c & p & s & Avg & c & p & r & s & Avg & a & c & d & w & Avg \\
        \hline
        CLIP-zs & 95.62& 97.23 & 99.40 & 80.33 & 93.15 & 80.38 & 77.48 & 90.28 &  74.70 & 80.71 & 96.91 & 89.91 & 94.44 & 93.65 & 93.73 \\
        \hline
        ResNet-full & 24.33 & 23.78 & 29.08 & 19.67 & 24.22 & 37.09 & 25.35 & 42.61 & 30.14 & 33.80 & 12.37 & 12.72 & 11.11 & 11.11 & 11.83 \\
        ViT-full  & 28.47 & 42.89 & 53.71 & 31.73 & 39.20 & 24.26 & 16.89 & 30.89 & 10.79 & 20.71 & 20.62 & 20.61 & 11.11 & 25.40 & 19.43 \\
        \hline
        ResNet-tuning & 75.43 & 64.75 & 80.71 & 62.56 & 70.86 & 80.54 & 74.55 & 88.57 & 76.79 & 80.11 & 56.19 & 52.19 & 50.00 & 52.38 & 52.69 \\
        ViT-tuning  & 87.83 & 88.11 & 97.33 & 38.58 & 77.96 & 84.21 & 77.24 & 89.82 & 78.12 & 82.35 & 86.60 & 83.33 & 88.89 & 82.54 & 85.34 \\
        CLIP-FC & 97.81 & 98.09 & 99.41 & 89.21 & 96.13 & 84.89 & 79.49 & 91.76 & 79.75 & 83.97 & 96.91 & 94.30 & 97.22 & 96.39 & 96.21 \\
        FedCLIP  & 97.81 & 97.03 & 99.70 & 90.86 & 96.35 & 83.96 & 80.42 & 92.36 & 79.07 & 83.95 & 96.91 & 94.74 & 97.22 & 95.24 & 96.03 \\
        FACMIC & 98.05 & 97.45 & 99.41 & 91.24 & 96.59 & 84.72 & 80.23 &  92.45 & 79.81 & 84.30 & 97.42 & 94.30 & 97.22 & 95.24 & 96.05 \\
        PromptFL  & 98.05 & 98.30 & 99.41 & 92.00 & 96.94 & 85.59 & 80.52 &  91.55 & 81.30 & 84.74 & 97.42 & \textbf{96.05} & 97.22 & 95.24 & 96.48 \\
        FedAPT & 98.05 & 98.51 & 99.41 & 92.13 & 97.03 & 85.45 & 80.37 & 92.46 & 82.19 & 85.12 & 97.42 & 94.74 & 100 & 96.83 & 97.25 \\
        \hline
        FedDEAP (Ours) & \textbf{98.54} & \textbf{99.15} & \textbf{99.70} & \textbf{98.86} & \textbf{99.06} & \textbf{86.45} & \textbf{82.33} & \textbf{92.78} & \textbf{83.50} & \textbf{86.27} & \textbf{97.94} & 95.18 & \textbf{100} & \textbf{98.41} & \textbf{97.88} \\
        \bottomrule
    \end{tabular}}
    \label{nat_results}
\end{table*}

\section{Experiments}
\subsection{Experimental Settings}
\textbf{Datasets.} 
We evaluated the proposed FedDEAP on three multi-domain natural image datasets including PACS \cite{li2017pacs}, DomainNet-126 \cite{peng2019domain}, and Office-Caltech10 \cite{gong2012office}. PACS consists of four visually distinct domains (Photo, Art Painting, Cartoon, and Sketch), each containing 7 shared categories. DomainNet-126 is a large-scale dataset with approximately 100,000 images from four domains (Clipart, Painting, Real, and Sketch), covering 126 object classes. Office-Caltech10 includes four domains (Amazon, Webcam, DSLR, and Caltech), all sharing 10 common categories, with variations in image acquisition conditions across domains.

To evaluate FedDEAP in medical image analysis, we used the DDR dataset \cite{li2019DDR}, which includes 13,673 fundus images labeled across five stages of diabetic retinopathy. To simulate domain shift, the data was divided into four domains based on resolution and illumination: HB (high resolution, bright), LB (low resolution, bright), HD (high resolution, dark), and LD (low resolution, dark).

% To evaluate the effectiveness of FedDEAP in medical image analysis scenarios, we employed a diabetic retinopathy classification dataset named DDR \cite{li2019DDR}. DDR contains 13,673 fundus images labeled according to five stages of disease severity from different hospitals. To simulate domain shift, we divided the dataset into four domains based on image resolution and illumination conditions: high resolution and bright (domain HB), low resolution and bright (domain LB), high resolution and dark (domain HD), and low resolution and dark (domain LD).

% \textbf{Client Partions.} We treated each domain in the dataset as the local data of a distinct client in a federated learning setting. For each client, we split their local dataset into a training set (80\%) and a test set (20\%). The clients collaboratively participated in federated training using their local training sets, and the aggregated global model was evaluated on each client's local test set to assess its generalization across domains.

\textbf{Compared Methods.} We compared FedDEAP with several competitive baselines. We began by evaluating the zero-shot classification capability of the CLIP model, where predictions are directly inferred using the pre-trained CLIP without any additional training. We then considered federated learning models that use \textbf{ResNet-50} \cite{he2016resnet} and \textbf{Vision Transformer} (ViT) \cite{dosovitskiy2020vit} as local backbones, aggregated via the FedAvg algorithm. For these models, we evaluated both training from scratch and fine-tuning of pre-trained backbones. The \textbf{CLIP-FC} baseline freezes all CLIP parameters and appends a trainable fully connected layer to the image encoder. \textbf{PromptFL} \cite{guo2023promptfl} fine-tunes client-specific prompts while aggregating them using FedAvg. \textbf{FedCLIP} \cite{lu2023fedclip} introduces a trainable adapter module after the image encoder, and performs federated aggregation on this adapter. \textbf{FACMIC} \cite{wu2024facmic} extends FedCLIP by incorporating a domain-level regularization term to enhance domain generalization. Lastly, \textbf{FedAPT} \cite{su2024fedapt} proposes a meta-prompt approach that incorporates client-specific information to generate personalized prompts conditioned on their data domains.

\begin{table}[h]
    \centering
    \caption{Comparison of classification accuracy between FedDEAP and baselines on DDR dataset.}
    \renewcommand{\arraystretch}{0.95}
    \begin{tabular}{l|ccccc}
        \toprule
        Method & HB & LB & HD & LD & Avg \\
        \hline
        ResNet-full & 60.92 & 57.88 & 60.86 & 63.92 & 60.90 \\
        ViT-full & 60.92 & 63.18 & 65.67 & 65.88 & 63.91 \\
        \hline
        ResNet-tuning & 71.37 & 67.99 & 72.47 & 73.73 & 71.39 \\
        ViT-tuning & 71.37 & 67.50 & 72.97 & 72.42 & 71.06 \\
        CLIP-FC & 71.50 & 71.48 & 74.79 & 78.95 & 74.18 \\
        FedCLIP & 55.29 & 52.07 & 60.36 & 59.35 & 56.77 \\
        FACMIC & 67.06 & 68.66 & 72.80 & 75.56 & 71.02 \\
        PromptFL & 72.42 & 70.15 & \textbf{75.62} & 79.08 & 74.32\\
        % FedLora & & & & & \\
        FedAPT & 73.07 & 69.15 & 75.46 & 78.82 & 74.12 \\
        \hline
        FedDEAP (Ours) & \textbf{74.25} & \textbf{71.97} & 74.79 & \textbf{80.78} & \textbf{75.45} \\
        \bottomrule
    \end{tabular}
    \label{medical_results}
\end{table}

\textbf{Implementation Details.} We conducted our experiments using the PyTorch library on an NVIDIA A100 GPU. We treated each domain in the dataset as the local data of a distinct client in a federated learning setting. For each client, we divide their local dataset into a training set (80\%) and a test set (20\%). Each client was assigned 16 tokens for both the personalized domain prompt and the global semantic prompt. For the PACS, Office, and DDR datasets, we set the learning rate to 0.001 and a batch size of 64. We trained the global model for 100 federated communication rounds. For the DomainNet dataset, we used a learning rate of 0.01 and a batch size of 256, training the global model for 50 rounds. The local training epoch per client was set to 1. All prompt tuning experiments were based on the pre-trained CLIP model with a ViT-B/32 backbone.

\subsection{Main Results}
Table 1 presents a comprehensive comparison between the proposed FedDEAP and several baselines on the PACS, DomainNet, and Office datasets. We summarize the following key observations: (1) The pre-trained CLIP model (CLIP-zs) exhibits strong zero-shot generalization capabilities across diverse natural image domains, as it effectively aligns image and text representations in a shared feature space. (2) Using pre-trained ResNet or Vision Transformer (ResNet-tuning/ViT-tuning) as backbone models in federated fine-tuning leads to better classification performance than training from scratch. (3) Our FedDEAP outperforms all baseline methods in nearly all domains and datasets, with the sole exception of the Caltech domain in the Office dataset. Notably, our method achieves a \textbf{6.73\%} improvement in the Sketch domain of PACS compared to the strongest baseline. In terms of average accuracy, we achieve \textbf{2.03\%} and \textbf{1.15\%} improvements over the best-performing baselines on PACS and DomainNet, respectively. (4) PromptFL demonstrates that federated prompt aggregation is more effective than adapter-based fine-tuning, such as FedCLIP and FACMIC. (5) FedAPT achieves the best performance among baselines by introducing client-specific key-value pairs.

% We further evaluated FedDEAP on the DDR retinal disease dataset. As shown in Table 2, FedDEAP achieves the highest accuracy in domains HB, LB, and LD, and obtains the best overall average accuracy across all four domains, outperforming the best baseline by 1.13\%. Compared to standard FedAvg using ResNet or ViT backbones (full/tuning), our FedDEAP demonstrates stronger adaptability and superior performance across various domains.
We further evaluated the performance of FedDEAP on the DDR retinal disease dataset. As presented in Table 2, FedDEAP achieves the highest classification accuracy in three out of the four domains—HB, LB, and LD. It also attains the best overall average accuracy across all domains, surpassing the strongest baseline by 1.13\%. In contrast to standard FedAvg approaches that use ResNet or ViT backbones, whether trained fully or with tuning, FedDEAP exhibits greater adaptability to domain shifts and consistently delivers superior performance in heterogeneous settings.

\subsection{Performance under Category Imbalance} 
% In federated learning, data heterogeneity arises not only from domain shift but also from imbalanced label distributions across clients. To evaluate the effectiveness of our proposed method under varying label distributions, we partitioned the data from each domain into three sub-datasets using a Dirichlet distribution. As a result, we obtained 12 clients, with each domain consisting of three clients. The concentration parameter $\alpha$ in Dirichlet distribution controls the degree of label distribution imbalance: smaller $\alpha$ values result in greater divergence in class distributions across sub-datasets. Following the experimental setup in \cite{su2024fedapt}, we simulated training by randomly selecting one client per domain in each round.

In federated learning, data heterogeneity arises not only from domain shift but also from imbalanced label distributions across clients. To evaluate the effectiveness of our proposed method under varying label distributions, we partitioned the data from each domain into three sub-datasets using a Dirichlet distribution. The concentration parameter $\alpha$ in Dirichlet distribution controls the degree of label distribution imbalance: smaller $\alpha$ values result in greater divergence in class distributions across sub-datasets.

We evaluated the performance of FedDEAP under varying degrees of label heterogeneity on PACS and DomainNet. Figure 2 compares the average classification accuracy across domains of our approach with four baseline methods. FedDEAP consistently outperforms all baselines across all $\alpha$ settings, demonstrating superior robustness in both highly heterogeneous and relatively balanced label distributions. Notably, under scenarios of extreme heterogeneity ($\alpha = 0.01, 0.1$), our FedDEAP achieves a substantial performance advantage over baseline methods, highlighting its effectiveness in mitigating performance degradation caused by severe client-side label heterogeneity.

\begin{figure}[htbp]
  \centering
  \begin{subfigure}[b]{0.49\linewidth}
    \centering
    \includegraphics[width=\linewidth]{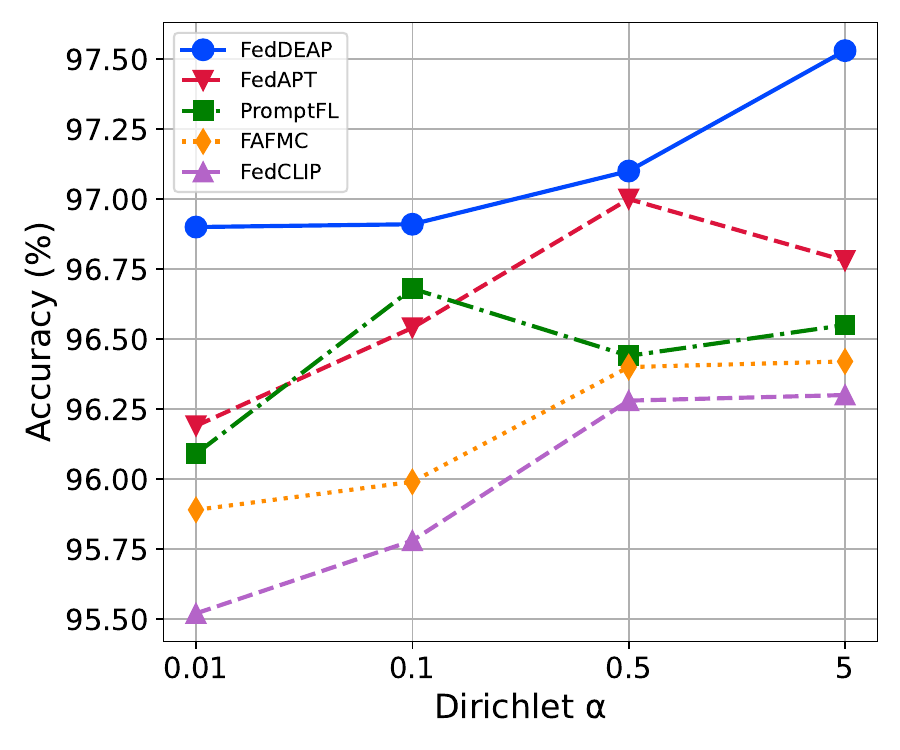}
    \caption{PACS}
    \label{non_iid_a}
  \end{subfigure}
  \hfill
  \begin{subfigure}[b]{0.49\linewidth}
    \centering
    \includegraphics[width=\linewidth]{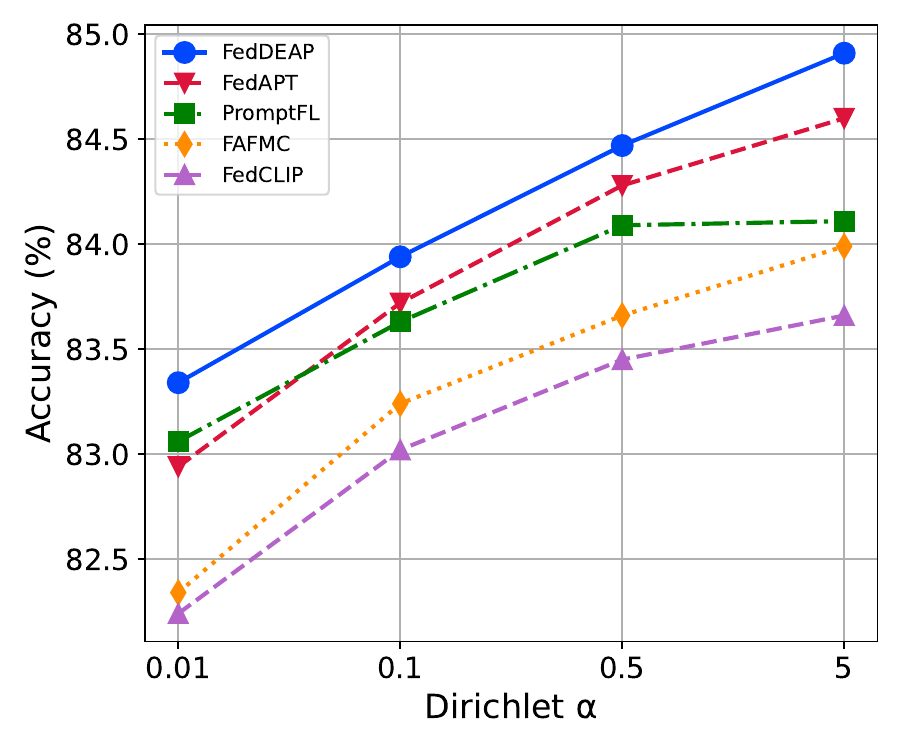}
    \caption{DomainNet}
    \label{non_iid_b}
  \end{subfigure}
  \caption{Average classification accuracy across domains under different Dirichlet $\alpha$ values on the PACS and DomainNet datasets.}
  \label{fig:strategy-comparison}
\end{figure}

\subsection{Domain Adaptivity of Prompts} 
To further validate whether the learned prompts possess robust domain-adaptive capabilities, we evaluated the matching effectiveness between prompts and image domains. Specifically, prompts trained on different image domains are applied to classify test images from all domains, and their accuracies across various combinations are compared. The experimental results are illustrated as heatmaps in Figure 3, with (a) and (b) representing results on the PACS and DomainNet datasets, respectively. As observed, the diagonal elements in both heatmaps exhibit the darkest colors, indicating that classification accuracy is highest when the image domain matches the prompt domain. This finding confirms that FedDEAP successfully learns prompts highly tailored to specific domains, substantially improving classification performance within corresponding domains.
% Moreover, for domains such as Sketch in both datasets, the accuracy gap between same-domain and cross-domain prompts is particularly notable, further demonstrating that our method effectively extracts domain-specific semantic information and achieves superior domain-awareness and adaptation capabilities.

\subsection{Prompt Embedding Visualization}
To further analyze the behavior of our proposed prompt-learning strategy in the feature space, we employed the t-SNE visualization method \cite{van2008tsne} to compare the distributions of textual features generated by prompts and test image features under different training strategies.

Figure 4 (a) and (b) show the distribution of textual features (stars) generated by the prompts trained using FedDEAP and PromptFL in the Cartoon domain of PACS, along with corresponding test image features (dots) and image cluster centroids (cross mark). It is evident that our textual prompt features closely align with the cluster centroids and exhibit clear separation between classes. However, textual features from the PromptFL's prompts exhibit overlap in some categories which indicates poor discrimination between classes. The experimental results demonstrate that our trained prompts accurately align with the semantic centers of different categories, demonstrating strong semantic discrimination and domain adaptability.

\begin{figure}[htbp]
  \centering
  \begin{subfigure}[b]{0.49\linewidth}
    \centering
    \includegraphics[width=\linewidth]{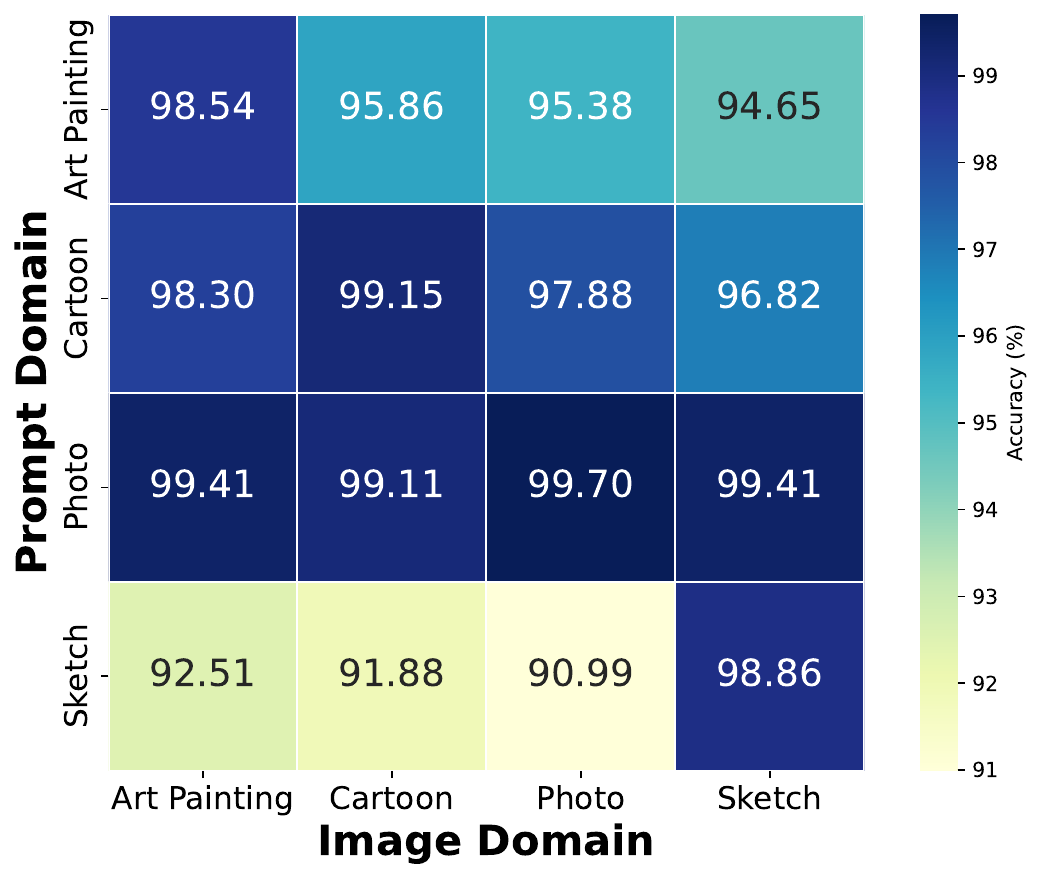}
    \caption{PACS}
    \label{adap_a}
  \end{subfigure}
  \hfill
  \begin{subfigure}[b]{0.49\linewidth}
    \centering
    \includegraphics[width=\linewidth]{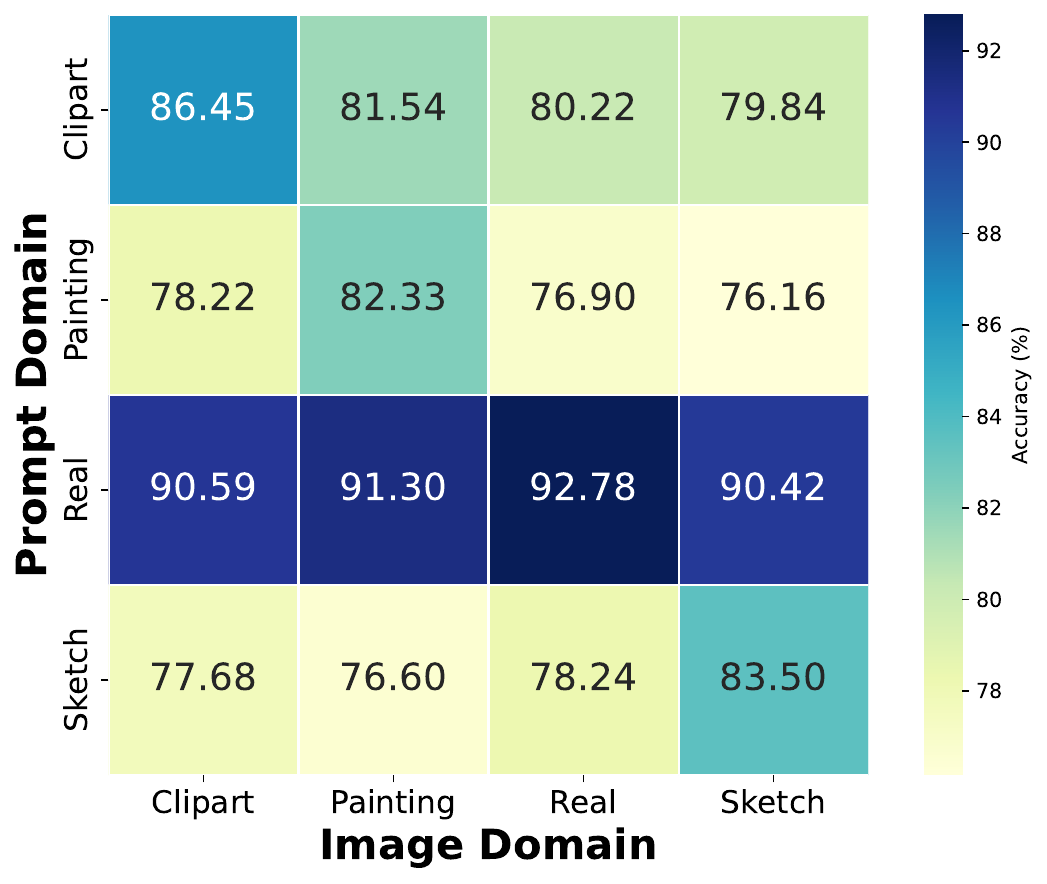}
    \caption{DomainNet}
    \label{adap_b}
  \end{subfigure}
  \caption{Classification accuracy heatmaps using prompts from different domains on various image domains in PACS and DomainNet.}
\end{figure}

\vspace{-2em}

\begin{table}[htbp]
\centering
\caption{Ablation study on different components across four datasets.}
\resizebox{\linewidth}{!}{
\begin{tabular}{lcccc}
\toprule
Component & PACS & DomainNet & Office & DDR \\
\midrule
Baseline & 96.94 & 84.74 & 96.48 & 74.32 \\
w/ Personalized Prompt & 97.57 & 85.32 & 97.08 & 74.80 \\
w/o Semantic Align.  & 98.67 & 85.98 & 97.49 & 75.06 \\
w/o Domain Align. & 98.56  & 86.10 & 97.25 & 75.16 \\
FedDEAP & \textbf{99.06} & \textbf{86.27} & \textbf{97.88} & \textbf{75.45} \\
\bottomrule
\end{tabular}}
\label{tab:ablation_study}
\end{table}

\vspace{-1em}

\begin{figure}[htbp]  % 使用 figure* 跨越双栏
  \centering
  \begin{subfigure}[b]{0.49\linewidth}
    \centering
    \includegraphics[width=\linewidth]{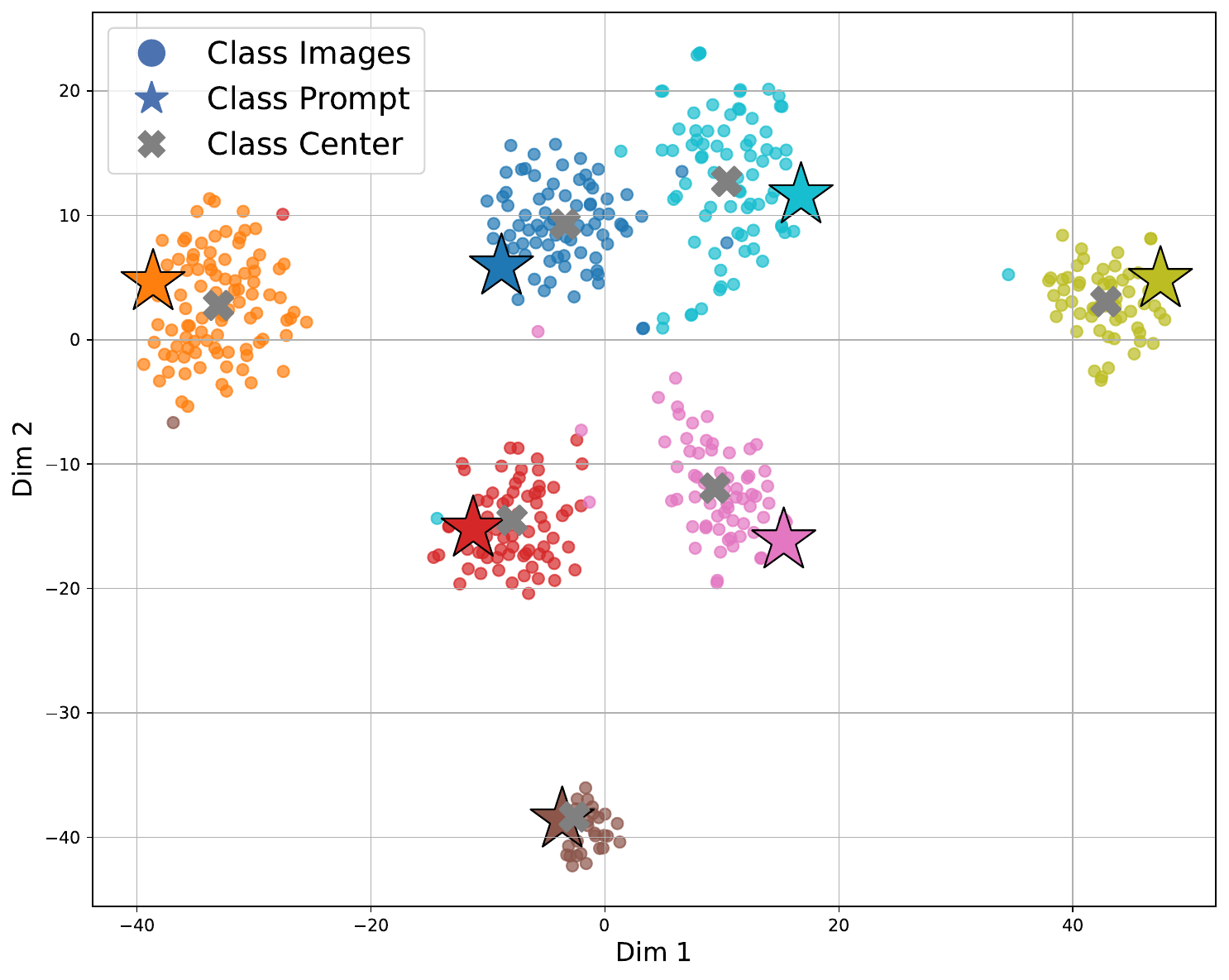}
    \caption{FedDEAP on PACS-Cartoon}
    \label{visual_a}
  \end{subfigure}
  \hfill
  \begin{subfigure}[b]{0.49\linewidth}
    \centering
    \includegraphics[width=\linewidth]{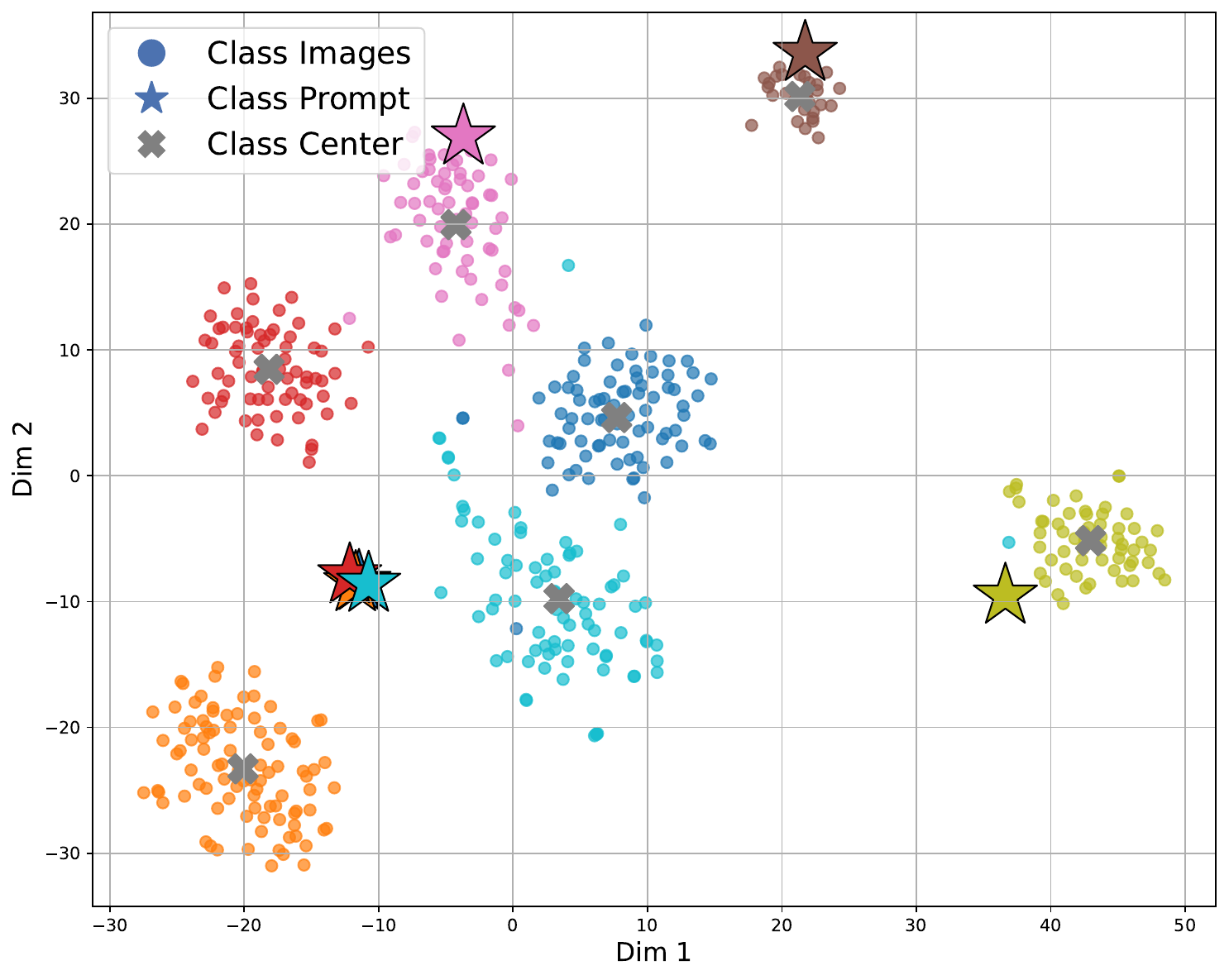}
    \caption{PromptFL on PACS-Cartoon}
    \label{fig:subfig-b}
  \end{subfigure}
   \caption{Visualization of prompt and image features in the embedding spaces of (a) FedDEAP and (b) PromptFL.}
  \label{visual_b}
\end{figure}

\subsection{Ablation Study}
To assess the effectiveness of each key component in our proposed method, we conducted an ablation study on four datasets. The results are summarized in Table 3. From the experimental results, we can derive some key insights: (1) Introducing personalized prompts significantly improves local adaptability. Compared to the baseline, adding personalized prompts consistently boosts performance across all datasets. This indicates that prompt representations learned by each client without aggregation can help capture the local structure of client-specific image distributions. (2) Removing the semantic alignment module leads to reduced global consistency, resulting in a noticeable drop in performance, especially on datasets with more pronounced label imbalance such as DomainNet. This demonstrates the effectiveness of semantic alignment in mitigating the impact of non-IID label distributions across clients. (3) Removing the domain alignment module impairs domain adaptability, leading to performance degradation on all datasets. This highlights the effectiveness of the domain alignment module in aligning prompt and image domains, thereby enhancing the ability of personalized prompts to capture domain-specific features. (4) Our full model achieves the highest performance with all the components. Compared to the baseline, our method achieves an absolute performance gain of \textbf{2.12\%}, \textbf{1.53\%}, \textbf{1.40\%}, and \textbf{1.13\%} on the four datasets, respectively. This confirms the complementary and synergistic contributions of different components in our method.

\begin{figure}[htbp]
  \centering
  \includegraphics[scale=0.29]{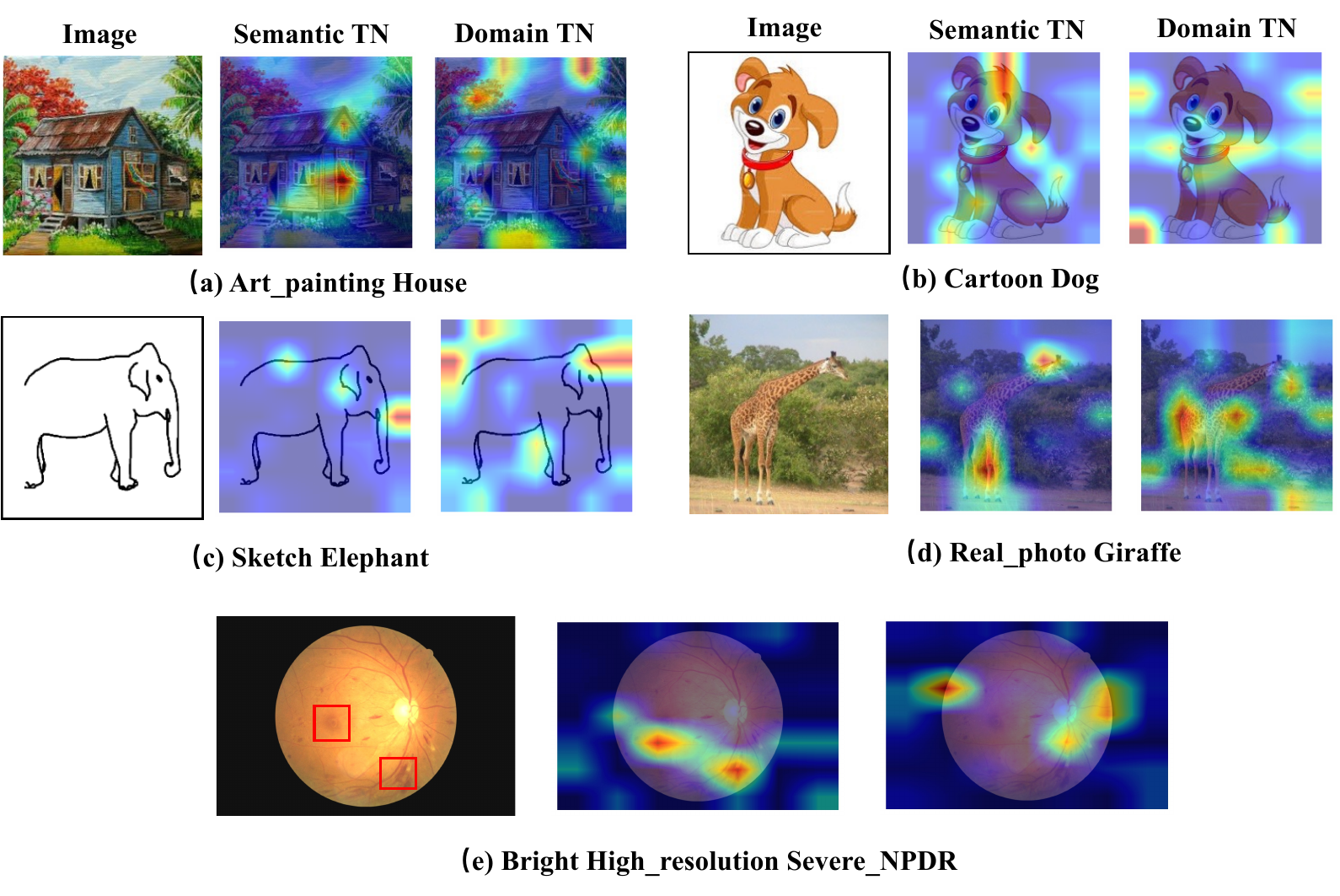}
  \caption{Grad-CAM analysis of semantic and domain transformation networks across different domains and categories.}
\end{figure}
\vspace{-10pt}

\subsection{Discussions and Limitations}
\textbf{Qualitative Analysis of Semantic and Domain Transformation Networks via Grad-CAM.} To better understand the behavior of the proposed semantic and domain transformation networks, we performed Grad-CAM \cite{selvaraju2017gradcam} on the outputs of the two transformation networks to visualize which regions in the input image contribute most to the transformed representations. Figure 5 illustrates the results on different domains and categories. Specifically, (e) shows a severe NPDR (non-proliferative diabetic retinopathy) sample from the DDR dataset, where the red bounding box indicates the lesion area. From the results of Figure 5 (a)-(e), we observe that the semantic transformation network focuses on semantically meaningful regions. Its activation maps are concentrated around object contours and edges, highlighting class-discriminative parts. In contrast, the domain transformation network captures global domain-specific patterns, such as background texture and drawing style, resulting in dispersed activations across the entire image.
\begin{table}[htbp]
\centering
\caption{Comparison of cost and performance across different methods}
\label{tab:cost_performance}
\resizebox{\linewidth}{!}{
\begin{tabular}{lcccc}
\toprule
Cost/Performance & FACMIC & PromptFL & FedAPT & FedDEAP \\
\midrule
Comm. (M/epoch) & 3.27 & \textbf{2.95} & 3.02 & 3.47 \\
Infer. Time (s/batch) & 2.89 & 2.54 & 6.43 & \textbf{2.54} \\
Performance (\%)  & 86.99 & 88.12 & 88.38 & \textbf{89.67} \\
\bottomrule
\end{tabular}}
\end{table}

\textbf{Efficiency Analysis.} We compared the efficiency of FedDEAP with other methods on the DomainNet dataset. As shown in Table 4, FedDEAP incurs a slightly higher communication cost per round compared to other baselines due to the upload of local semantic and domain transformation networks. However, FedDEAP achieves significantly faster inference speed compared to the best-performing baseline FedAPT. Moreover, it attains the highest average classification accuracy across four datasets, reaching 89.67\%. This demonstrates that FedDEAP offers a better trade-off between efficiency and effectiveness, achieving superior accuracy with minimal inference overhead.

\begin{table}[htbp]
\centering
\caption{Effect of Number of Personalized and Global Prompts on DomainNet}
\label{tab:prompt_config}
\resizebox{\linewidth}{!}{
\begin{tabular}{lccccc}
\toprule
Num. of Tokens & c & p & r & s & Avg \\
\midrule
p\_prom.=12, g\_prom.=20 & \textbf{86.61} & 81.84 & 92.55 & \textbf{83.64} & 86.16 \\
p\_prom.=20, g\_prom.=12 & 86.29 & 81.30 & 92.76 & 83.54 & 85.97 \\
p\_prom.=16, g\_prom.=16 & 86.45 & \textbf{82.33} & \textbf{92.78} & 83.50 & \textbf{86.27} \\
\bottomrule
\end{tabular}}
\end{table}

\textbf{Effect of Prompt Ratio.} We conducted a study to investigate how the ratio of personalized and global prompts influences model performance. As shown in Table 5, increasing the number of personalized prompts (from 12 to 16) enables the model to better capture domain-specific features, which improves performance on individual domains. However, reducing the number of global prompts weakens the model’s ability to generalize across domains, leading to a drop in average accuracy. The best overall performance is achieved with a balanced configuration, indicating that an appropriate allocation between personalized and global prompts is essential for achieving both effective domain adaptation and robust cross-domain generalization.

\section{Conclusion}
In this paper, we propose a federated prompt fine-tuning approach FedDEAP for the CLIP model to enhance cross-domain image recognition performance. Our method integrates global prompts with personalized local prompts, enabling adaptation to individual data domains while preserving global semantic knowledge. Specifically, we transform images from each domain into unbiased semantic and domain feature spaces. The unbiased semantic and domain transformation networks are trained and utilized to align prompts and images in both semantic and domain feature spaces, effectively encoding global semantic representations and local domain-specific characteristics into prompts. Our approach effectively addresses challenges posed by domain shift and class heterogeneity in federated learning, achieving improved image classification performance across multiple domains on several benchmark datasets.

\section{Acknowledgment}
This work was supported in part by National Natural Science Foundation of China (Grant No. 62441227), National Key Research and Development Program of China (Grant No. 2023YFB3106500), and China Scholarship Council (Grant No. 202406230318). Pak-Hei Yeung is grateful for support from the Presidential Postdoctoral Fellowship by the Nanyang Technological University.

\bibliographystyle{ACM-Reference-Format}
\bibliography{sample-base}
\vspace{15 pt}
\appendix
\begin{center}
    {\LARGE \textbf{Appendix}}  % 字号可改为 \Huge / \Large
\end{center}
\section{Bounded Distance Between Transformed Text and Image Representations of the Same Class}

Due to the equiangular tight property of the ETF, the cosine similarity between any two prototypes of an ETF with $K$ classes is defined as:
\begin{equation}
    \cos\theta = -\frac{1}{K-1}
\end{equation}

For prompt feature representations, their transformed outputs are linearly aligned with the ETF prototypes. After sufficient training and optimization, the angular distance between the prompt feature and the ETF prototype belonging to the same class becomes very small. Consequently, within the same class, the angle between the image feature $r_i$ and the prompt feature $r_t$ after transformation is highly likely to be smaller than $\frac{\theta}{2}$. Based on this observation, we can derive that:
\begin{equation}
    P\left(\cos\left(\Phi_s(r_t), \Phi_s(r_i)\right) \geq \cos\frac{\theta}{2}\right) \to 1
\end{equation} 

After normalizing both transformed vectors, the Euclidean distance can be bounded as:
\begin{equation}
\begin{aligned}
    \|\Phi_s(r_t) - \Phi_s(r_i)\| 
    &\leq \sqrt{\|\Phi_s(r_t)\|^2 + \|\Phi_s(r_i)\|^2 - 2\langle\Phi_s(r_t), \Phi_s(r_i)\rangle} \\
    &= \sqrt{2 - 2\cos(\Phi_s(r_t), \Phi_s(r_i))} \\
    &\leq \sqrt{2 - 2\cos\frac{\theta}{2}}
\end{aligned}
\end{equation}

Given $\cos\theta = -\frac{1}{K-1}$, we compute:
\begin{equation}
    \cos\frac{\theta}{2}
    = \sqrt{\frac{1 + \cos\theta}{2}} 
    = \sqrt{\frac{1 - \frac{1}{K-1}}{2}} 
    = \sqrt{\frac{K - 2}{2(K - 1)}}
\end{equation}

Thus, the upper bound of the distance becomes:
\begin{equation}
\begin{aligned}
    \|\Phi_s(v_1) - \Phi_s(v_2)\| 
    &\leq \sqrt{2 \left(1 - \sqrt{\frac{K - 2}{2(K - 1)}}\right)} \\
    &\leq \sqrt{2 - \sqrt{\frac{2K - 4}{K - 1}}}
\end{aligned}
\end{equation}

Therefore, for class $k$, we can conclude:
\begin{equation}
    P\left(\|\Phi_s(r_t)-\Phi_s(r_i)\| \leq \delta \mid k\right) \geq \gamma
\end{equation}
when $\delta = \sqrt{2 - \sqrt{\frac{2K - 4}{K - 1}}}$, $\gamma \to 1$.

In summary, the ETF structure enforces uniform angular separation among class prototypes and implicitly regularizes the feature geometry. As a result, the transformed prompt and image features within the same class converge to a compact region bounded by $\delta$, providing a theoretical guarantee for intra-class consistency in the learned representation space.

\section{Lower Bound of Conditional Entropy}

When the prompt feature representation is exactly aligned with its corresponding ETF prototype, 
the cosine similarity between the prompt feature and the prototype reaches its maximum possible value of $1$. 
In this ideal case, the transformed prompt feature $\Phi_s(r_t)$ perfectly matches the prototype of class $k$, 
leading to an extremely confident prediction for that class and negligible probabilities assigned to all others. 
As a result, the conditional entropy of the prediction distribution reaches its minimum value.

To analyze this more concretely, we examine the logit structure implied by the ETF geometry. 
When the ETF prototypes form equal angular separations, the similarity-based logits can be expressed as:
\begin{itemize}
    \item The logit for the correct class (e.g., class 1) is given by
    \begin{equation}
    l_1 = 1
    \end{equation}
    since $\cos(\Phi_s(r_t), v_1) = 1$ under perfect alignment.
    \item For each of the remaining $K - 1$ incorrect classes, the equiangular tight frame property ensures
    \begin{equation}
    l_i = -\frac{1}{K - 1}, \quad \text{for } i = 2, 3, \ldots, K
    \end{equation}
    reflecting the uniform angular separation between distinct prototypes.
\end{itemize}

\noindent
\textbf{Softmax probabilities.} 
Under the standard Softmax formulation, the predicted class probabilities are obtained by exponentiating and normalizing the logits:
\begin{equation}
p_i = \frac{e^{l_i}}{\sum_{j=1}^{K} e^{l_j}}
\end{equation}
Substituting the above logit values, we obtain:
\begin{itemize}
    \item For the correct class:
    \begin{equation}
    p_1 = \frac{e^1}{e^1 + (K - 1)e^{-1/(K - 1)}}
    \end{equation}
    \item For each incorrect class:
    \begin{equation}
    p_i = \frac{e^{-1/(K - 1)}}{e^1 + (K - 1)e^{-1/(K - 1)}}, \quad i = 2, \ldots, K
    \end{equation}
\end{itemize}

\noindent
Intuitively, this Softmax structure captures the confidence concentration effect: 
as the prompt feature aligns more closely with its prototype, the correct-class logit dominates exponentially, 
pushing $p_1$ toward $1$ while shrinking all other $p_i$ toward $0$. 
To simplify notation, we define the normalization constant:
\begin{equation}
Z = e + (K - 1)e^{-1/(K - 1)}
\end{equation}
Then the class probabilities can be compactly written as:
\begin{equation}
p_1 = \frac{e}{Z}, \qquad p_i = \frac{e^{-1/(K - 1)}}{Z}
\end{equation}

\noindent
\textbf{Entropy derivation.} 
The conditional entropy of this probability distribution is:
\begin{equation}
H = -\sum_{i=1}^{K} p_i \log p_i
\end{equation}
Substituting the probabilities yields:
\begin{equation}
H(K) = -\left[\frac{e}{Z} \log\!\left(\frac{e}{Z}\right) 
+ (K - 1)\frac{e^{-1/(K - 1)}}{Z} 
\log\!\left(\frac{e^{-1/(K - 1)}}{Z}\right)\right]
\end{equation}
After simplification, the closed-form entropy as a function of $K$ becomes:
\begin{equation}
H(K) = \log\!\left[e + (K - 1)e^{-1/(K - 1)}\right]
- \frac{e - e^{-1/(K - 1)}}{e + (K - 1)e^{-1/(K - 1)}}
\end{equation}

\noindent
This value corresponds to the minimum conditional entropy achievable when the transformed prompt feature is perfectly aligned with its class-$k$ ETF prototype. It represents the theoretical lower bound on classification uncertainty under ideal alignment, 
serving as a baseline for evaluating how far a learned representation deviates from the optimal geometric configuration.

In essence, this derivation connects the geometric regularity of the ETF structure 
with the information-theoretic behavior of the classifier: 
as the feature--prototype alignment improves, the softmax distribution becomes increasingly peaked, 
thereby minimizing entropy and maximizing classification confidence.

\section{Additional Experiments}
To investigate the influence of the hyperparameters $\lambda$ and $\eta$ on model performance, we conducted experiments on four benchmark datasets: PACS, Office, DomainNet, and DDR. The results are summarized in Figure 6. 

From the experimental results, it can be observed that when both $\lambda$ and $\eta$ are set to 1, the model achieves competitive performance across all four datasets. This indicates that increasing the weighting of the alignment loss between the prompt features and the ETF enhances the global consistency of semantic prompts while improving the adaptability of domain-specific prompts to different domains, thereby leading to superior classification performance.

\begin{figure}[h]
  \centering
  \includegraphics[scale=0.35]{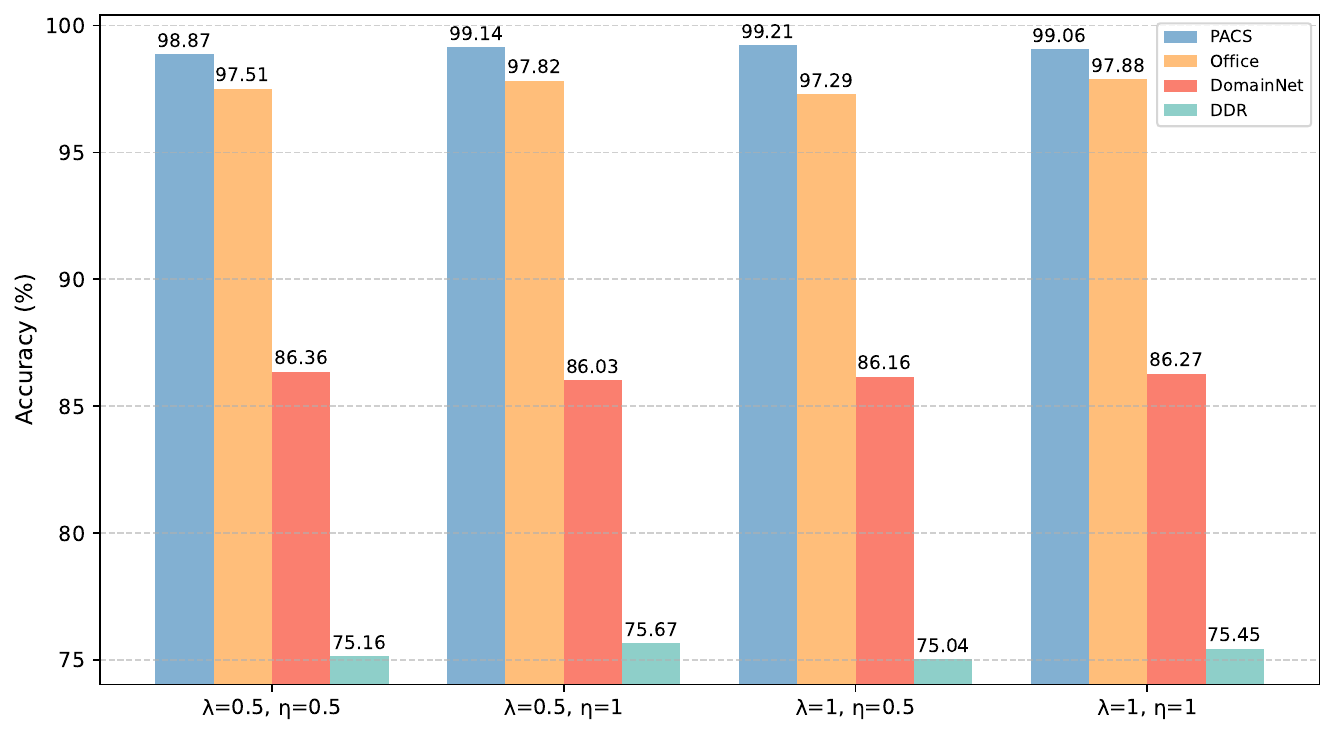}
  \caption{Accuracy comparison under different hyperparameter settings of $\lambda$ and $\eta$ on four datasets.}
\end{figure}

\begin{figure}[htbp]  % 使用 figure* 跨越双栏
  \centering
  \begin{subfigure}[b]{0.49\linewidth}
    \centering
    \includegraphics[width=\linewidth]{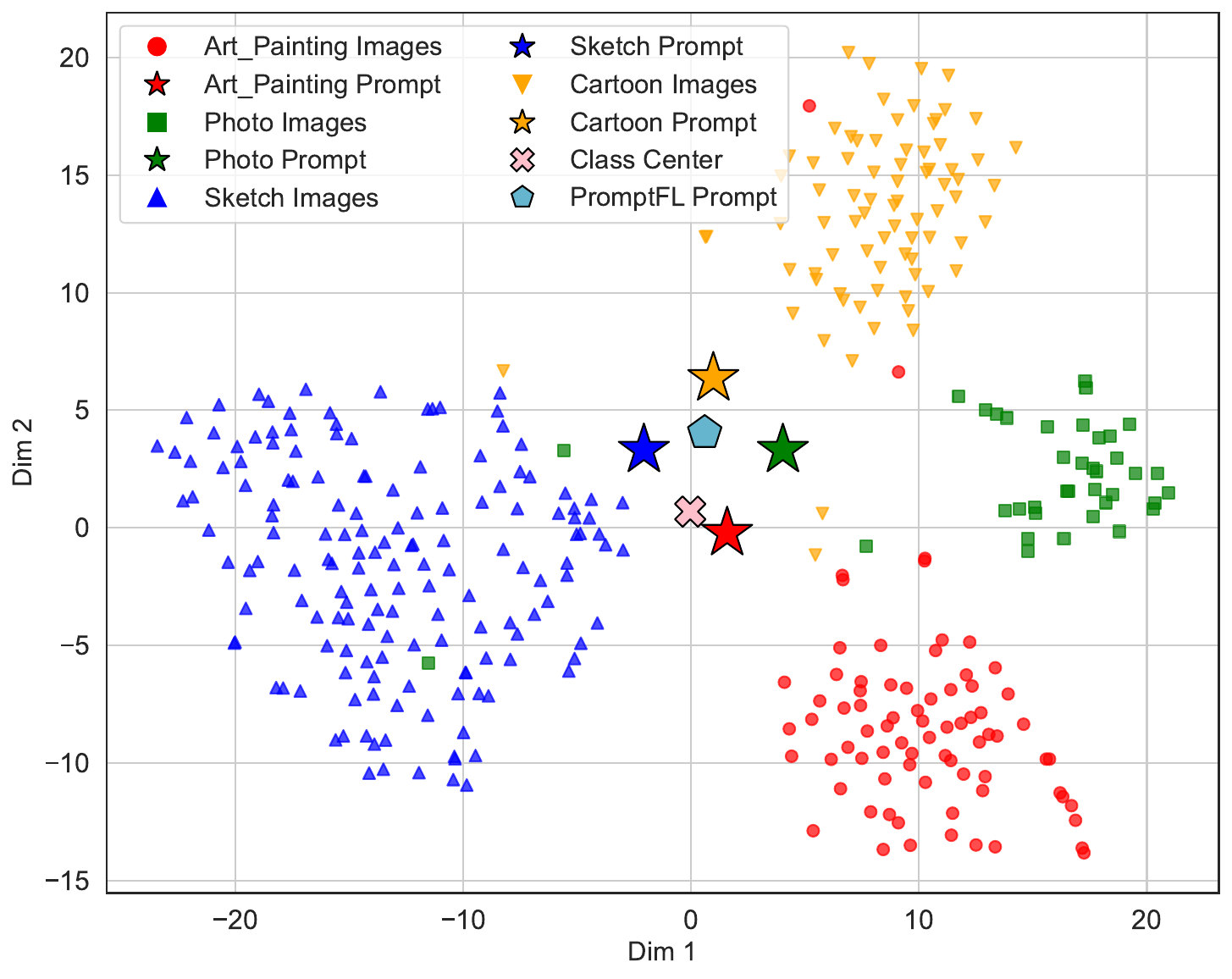}
    \caption{PACS}
    \label{visual_a}
  \end{subfigure}
  \hfill
  \begin{subfigure}[b]{0.49\linewidth}
    \centering
    \includegraphics[width=\linewidth]{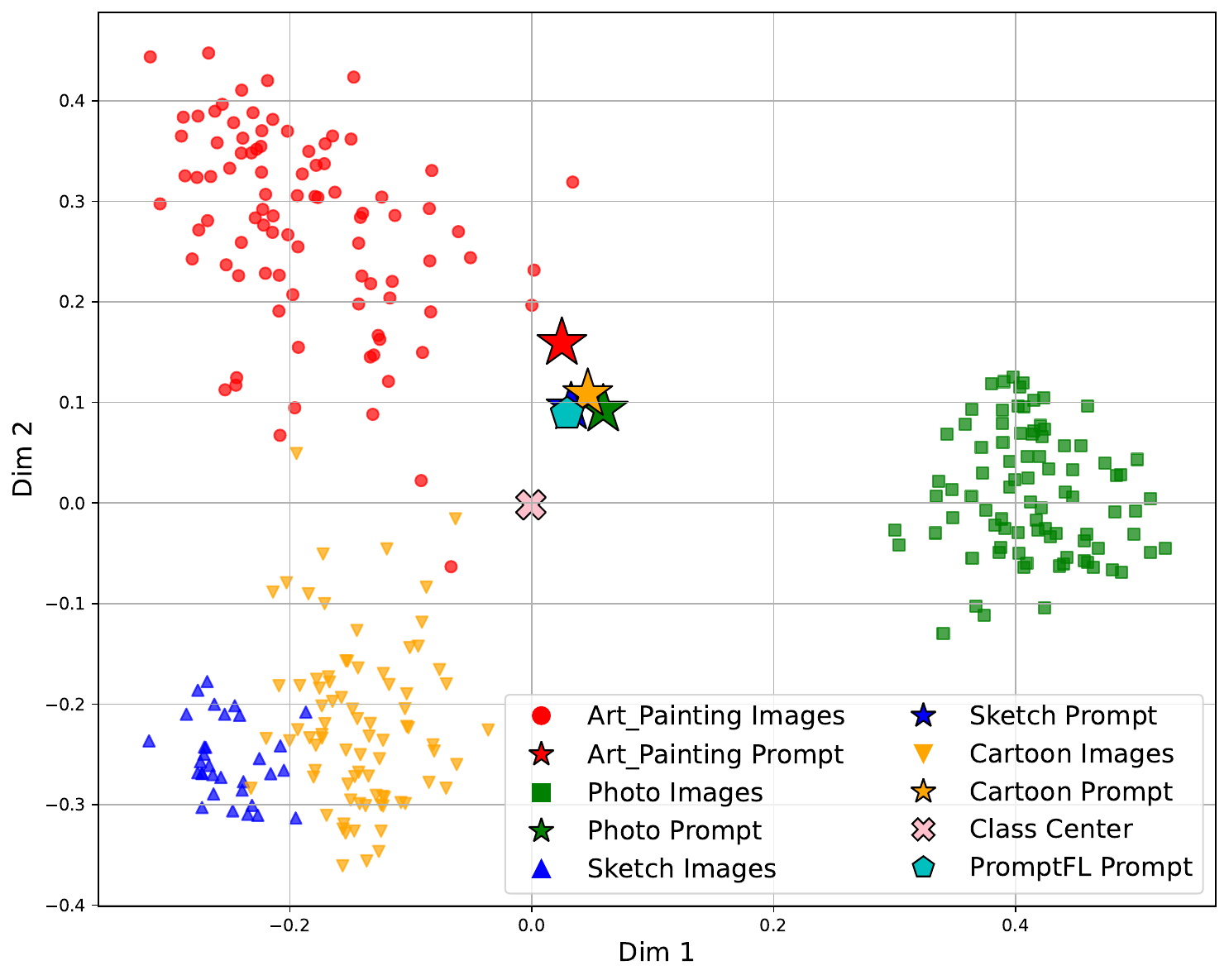}
    \caption{DomainNet}
    \label{fig:subfig-b}
  \end{subfigure}
   \caption{Cross-domain visualization of prompts and images from different domains for the same class in (a) PACS and (b) DomainNet.}
  \label{visual_b}
\end{figure}

Figure 7 provides a deeper analysis of the prompt feature distributions learned by our method across multiple domains for the same object category. Specifically, we employ t-SNE to project both image and prompt features into a two-dimensional space for intuitive comparison. The visualization reveals that the prompts corresponding to the same category, although originating from different domains, not only cluster closely with their respective domain-specific image features but also converge toward a shared semantic center. This behavior demonstrates that our method effectively captures domain-invariant semantics while preserving domain-specific characteristics, thereby achieving a balance between global consistency and local adaptability. 

Additionally, compared to PromptFL's global prompt, our prompt feature distribution more closely reflects the true image distributions across each domain. The results indicate that our method effectively learns discriminative semantic representations and integrates domain-specific information, achieving strong semantic alignment and generalization on heterogeneous multi-domain data.

\end{document}